\documentclass[sn-mathphys-num]{sn-jnl}%

\usepackage{subfigure}
\usepackage{hyperref}
\usepackage{comment}
\usepackage{siunitx}
\usepackage{fancyhdr}
\usepackage{enumerate}
\usepackage[utf8]{inputenc}
\usepackage{array}
\usepackage{longtable}

\usepackage{multirow}%
\usepackage{amsmath,amssymb,amsfonts}%
\usepackage{amsthm}%
\usepackage{mathrsfs}%
\usepackage[title]{appendix}%
\usepackage{xcolor}%
\usepackage{textcomp}%
\usepackage{manyfoot}%
\usepackage{booktabs}%
\usepackage{algorithm}%
\usepackage{algorithmicx}%
\usepackage{algpseudocode}%
\usepackage{listings}%

\theoremstyle{thmstyleone}%

\theoremstyle{thmstyletwo}%

\theoremstyle{thmstylethree}%

\begin{document}

\title[Medical Image Registration and Its Application in Retinal Images: A Review]{Medical Image Registration and Its Application in Retinal Images: A Review}

\author[1]{\fnm{Qiushi} \sur{Nie}}\email{12232413@mail.sustech.edu.cn}
\author[1,2]{\fnm{Xiaoqing} \sur{Zhang}}\email{xq.zhang2@siat.ac.cn}
\author[1]{\fnm{Yan} \sur{Hu}}\email{huy3@sustech.edu.cn}
\author[1]{\fnm{Mingdao} \sur{Gong}}\email{12011204@mail.sustech.edu.cn}
\author*[1,3,4,5]{\fnm{Jiang} \sur{Liu}\email{liuj@sustech.edu.cn}}

\affil[1]{\orgdiv{Research Institute of Trustworthy Autonomous Systems and Department of Computer Science and Engineering}, \orgname{Southern University of Science and Technology}, \orgaddress{\city{Shenzhen} \postcode{518055}, \country{China}}}

\affil[2]{\orgdiv{Center for High Performance Computing and Shenzhen Key Laboratory of Intelligent Bioinformatics, Shenzhen institute of Advanced Technology}, \orgname{Chinese Academy of Sciences}, \orgaddress{\city{Shenzhen} \postcode{518055}, \country{China}}}

\affil[3]{\orgdiv{Research Institute of Trustworthy Autonomous Systems and Department of Computer Science and Engineering}, \orgname{Southern University of Science and Technology}, \orgaddress{\city{Shenzhen} \postcode{518055}, \country{China}}}

\affil[4]{\orgdiv{Singapore Eye Research Institute}, \orgaddress{\postcode{169856},  \country{Singapore}}}

\affil[5]{\orgdiv{State Key Laboratory of Ophthalmology, Optometry and Visual Science, Eye Hospital}, \orgname{Wenzhou Medical University}, \orgaddress{\city{Wenzhou} \postcode{325027}, \country{China}}}

\abstract{Medical image registration is vital for disease diagnosis and treatment with its ability to merge diverse information of images, which may be captured under different times, angles, or modalities. Although several surveys have reviewed the development of medical image registration, these surveys have not systematically summarized methodologies of existing medical image registration methods. To this end, we provide a comprehensive review of these methods from traditional and deep learning-based directions, aiming to help audiences understand the development of medical image registration quickly. In particular, we review recent advances in retinal image registration at the end of each section, which has not attracted much attention. Additionally, we also discuss the current challenges of retinal image registration and provide insights and prospects for future research.}


\maketitle

\section{Introduction}
\label{sec:introduction}

Medical image registration is a fundamental step in computer-aided diagnosis (CAD) and image-guided surgical treatment, attracting much attention. It aligns multiple medical images by finding appropriate spatial transformation relationships for fusing their corresponding information, helping doctors make a more comprehensive and precise diagnosis conclusion. Particularly, these medical images may be acquired at different times, angles, and even modalities of a certain tissue or organ of the human body as input. Therefore, the nature of medical image registration is to eliminate the interference of these factors and find consistent objects or shapes for matching.

To deal with different transformation tasks in medical image registration, massive methods have been developed. They can be grouped into two types: coarse-grained global linear registration and fine-grained local elastic registration. Coarse-grained global linear registration extracts salient features of the input image pair, thereby matching these features and overcoming angular changes. Fine-grained local elastic registration performs pixel-level analysis of the input image pair after linear alignment and performs local corrections to overcome spontaneous tissue movements and deformations.

Another way to classify registration methods is according to what is used to match the images. A first and direct approach is intensity-based methods \cite{oliveira2014medical}. These methods consider registration as an optimization problem by iteratively disturbing the transformation parameters to maximize the pixel-wise similarity. Another early but still popular approach is feature-based methods \cite{zitova2003image}, which extract manually designed features and descriptors, match them, and establish the transformation based on the matching. In contrast to intensity-based methods, feature-based methods provide more robust registration by matching salient features rather than simply comparing the pixels. 

In the past decade, deep features have taken the place of handcraft features with their ability to provide learnable and, therefore, more flexible and problem-specific feature representations for registration tasks. Later, after the deep feature extractors, end-to-end registration neural networks integrated the whole registration process into a single network, applying deep-learning techniques such as convolutional neural networks (CNN), generative adversarial networks (GAN), and Transformers. Once trained, these methods can obtain registration results directly from input image pairs, thereby speeding up registration, and they have also been proven to have better registration performances. 

Several reviews have been conducted on deep learning for medical image registration \cite{boveiri2020medical, haskins2020deep, bharati2022deep}. However, these papers only investigated the popular CNN-based methods at the time, which did not mention the latest Transformer-based methods. Additionally, these works only investigated methods based on deep learning, but ignored traditional methods from the early years, which can also provide significant guidance. 

Among medical images, retinal images focus on a unique part of the human body that allows for non-invasive observation of blood vessels in vivo. Retinal image registration, which combines complementary structural and functional information from the same or different modalities, is a crucial step in the process. However, in the past few years, few surveys have systematically reviewed retinal image registration. Saha \emph{et al.} \cite{saha2019color} and Pan \emph{et al.} \cite{pan2021retinal} reviewed retinal image registration but focused solely on registering one specific retinal modality. Moreover, they did not compare these methods with mainstream medical image registration methods.

The purpose of this paper is to review and summarize the existing medical image registration works from traditional methods and deep learning-based methods, aiming to help audiences grasp the development of medical image registration clearly. Moreover, we also survey and synthesize retinal image registration works as a particular characteristic of this review. Finally, we also highlight the current challenges of retinal image registration and discuss future research directions.

\begin{figure*}[htbp]
    \centering
    \includegraphics[width=\linewidth]{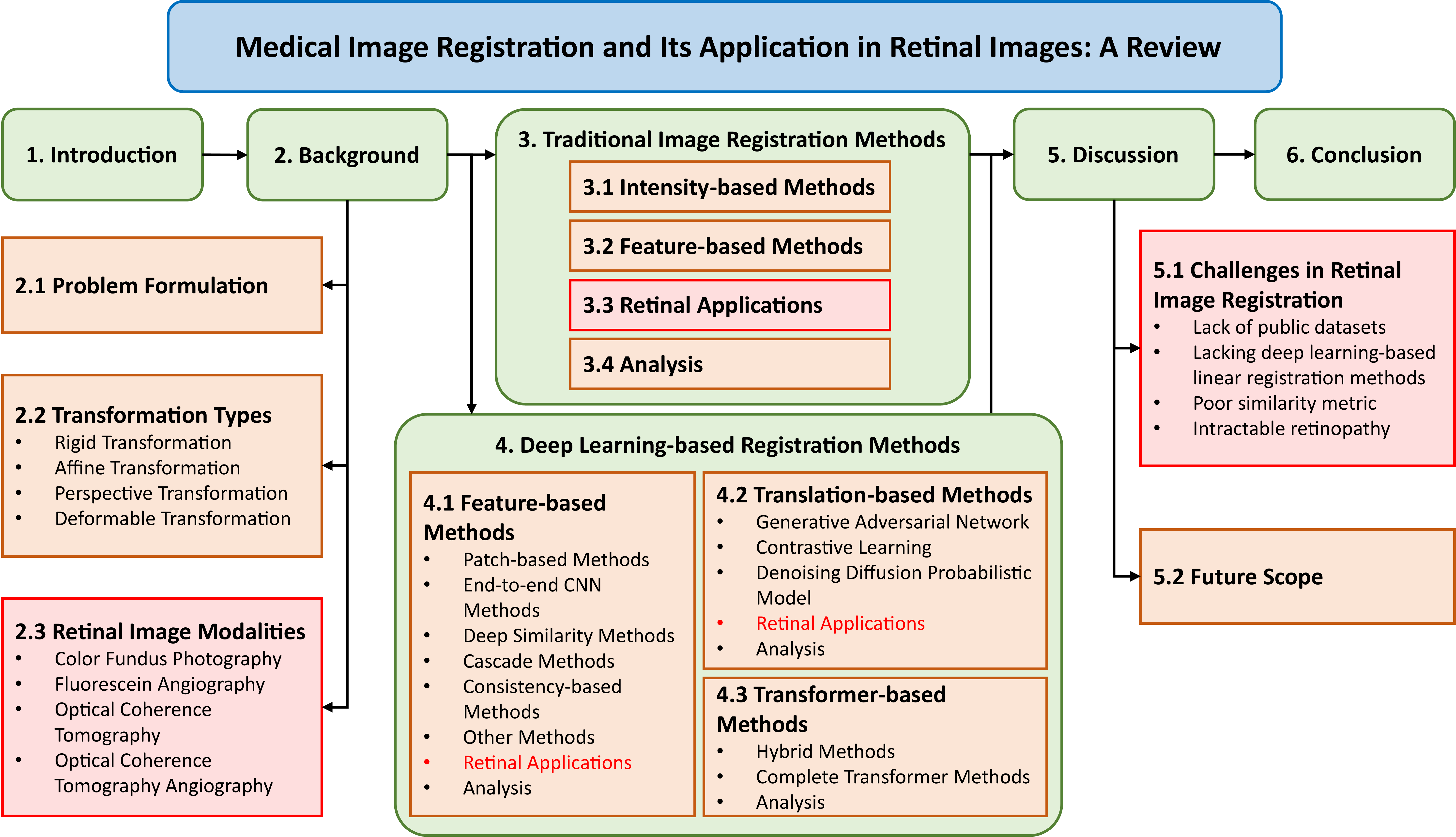}
    \caption{Structure of our review}
    \label{contents}
\end{figure*}

The overall organization is presented in Fig. \ref{contents}: Section \ref{sec:bg} defines the basic concepts of image registration and briefly introduces the popular retinal image modalities. Section \ref{sec:mlm} and Section \ref{sec:dlm} review the general methodology of medical image registration, the corresponding applications in retinal image registration, and comparisons between them, categorized by traditional and deep learning, respectively. Furthermore, Section \ref{sec:discussion} discusses the advantages and disadvantages of the reviewed methods, points out the current challenges, and provides potential future research directions. Finally, in Section \ref{sec:conclusion}, we summarize the paper.

\section{Background}
\label{sec:bg}

\subsection{Problem Formulation}

Image registration is a fundamental task in image processing. It involves finding correspondences between two images, namely a moving and a fixed image, and establishing a transformation between them. The fixed image is used as a reference, and the goal is to transform the moving image to match the fixed image. Registration algorithms are designed to find the best transformation, denoted by $T^*$, that maximizes the similarity between the two images \cite{khalifa2011state}. This can be achieved by maximizing the image similarity function $\text{sim}(I_f, T(I_m))$, where $I_m$ and $I_f$ are the moving and fixed images, respectively, and $T(I_m)$ is the transformed moving image using the transform $T$.

\subsection{Transformation Types}

\begin{figure}[htbp]
	\centering
	\vspace{-0.5cm}
	\subfigure[Origin]{\includegraphics[scale=0.57]{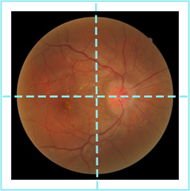}}
	\subfigure[Rigid]{\includegraphics[scale=0.57]{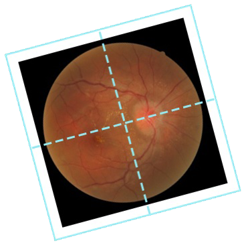}}
	\subfigure[Affine]{\includegraphics[scale=0.57]{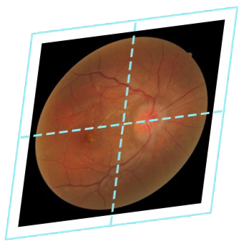}}
	\vspace{1cm}
	\subfigure[Perspective]{\includegraphics[scale=0.57]{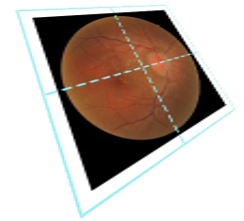}}
	\subfigure[Deformable]{\includegraphics[scale=0.58]{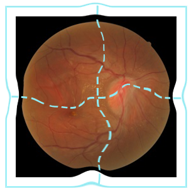}}
    \caption{The effect of different transformations}
    \label{transforms}
\end{figure}

In this section, we introduce different transformation models. There are four main types of transformation: rigid, affine, perspective, and deformable. The first three are linear transformations capable of executing macro adjustments, while the fourth is non-linear and corrects local discrepancies. Fig. \ref{transforms} offers a visual representation of the effects of these transformations. Although the descriptions and figures below pertain to the 2D registration, inferences can be drawn for 3D registration.

\subsubsection{Rigid Transformation}
\emph{Rigid transformation} is a widely used simple and fast method in image processing. It consists of translation and rotation and never changes the size or shape of the original image. The model of rigid body transformation can be represented as:
\begin{equation}
    \label{affine_trans}
    \begin{bmatrix}x'\\y'\end{bmatrix}
    = \boldsymbol{R}\begin{bmatrix}x\\y\end{bmatrix}+\boldsymbol{t}
\end{equation}
Here, $(x, y)$ is the coordinate of the pixel in the image to be transformed, and $(x', y')$ is the target pixel in the transformed image, $\boldsymbol{R}=\begin{bmatrix} \cos\theta & -\sin\theta\\ \sin\theta & \cos\theta\\ \end{bmatrix}$ is the rotation matrix, and $\boldsymbol{t} = [t_x, t_y]^T$ is the translation vector.

\subsubsection{Affine Transformation}
\emph{Affine transformation} is realized by combining a series of atomic transformations. Based on rigid transformation, affine transformation adds scaling and shearing. The model of affine transformation can be represented as:
\begin{equation}
    \label{rigid_trans}
    \begin{bmatrix}x'\\y'\end{bmatrix}
    = \begin{bmatrix}
        a & b\\c & d
    \end{bmatrix}\begin{bmatrix}x\\y\end{bmatrix}+
    \boldsymbol{t}
\end{equation}
Compared to rigid transformation, affine transformation adds more freedom to the rotation matrix to handle more significant differences between images than rigid transformation. It can map straight lines to straight lines and retains the property of preserving parallelism. It maintains the parallel relationship between lines but cannot maintain the vertical relationship between lines.

\subsubsection{Perspective Transformation}

\emph{Perspective transformation}, or projective transformation, is a more advanced form of transformation that corrects perspective distortions between images. Perspective transformation can correct more complex distortions such as foreshortening, skew, and non-parallelism. 

Perspective transformation involves finding a transform matrix in homogeneous coordinates. The homogeneous coordinates use a tuple of 3 numbers $(x_h,y_h,w_h)$ as point representation and can be translated from Cartesian coordinates $(x_c,y_c)$ by any $w_h\in\mathcal{R}$, $x_h=w_hx_c,\ y_h=w_hy_c$. In homogeneous coordinates, the transformation can be defined as:
\begin{equation}
    \label{pers_trans_in_homogeneous}
    \begin{bmatrix}x'\\y'\\w'\end{bmatrix}
    = \begin{bmatrix}
        A & B & C\\D & E & F\\a & b & c
    \end{bmatrix}
    \begin{bmatrix}x\\y\\w\end{bmatrix}
\end{equation}
Here, $(x, y, w)$ is the image's homogeneous coordinate to be transformed, $(x', y', w')$ is the target coordinate in the transformed image. By setting $w=1$ and transform the target $w'=1$, we have the target point $(x', y')$ back in Cartesian coordinate
\begin{equation}
    \label{pers_trans_in_homogeneous}
    \begin{aligned}
    x' &= \frac{Ax+By+C}{ax+by+c}\\
    y' &= \frac{Dx+Ey+F}{ax+by+c}\\
    \end{aligned}
\end{equation}
The perspective transformation has a straightness-preserving property, which means that a straight line is mapped to a straight line, but neither parallelism nor perpendicularity can be guaranteed.

\subsubsection{Deformable Transformation}
\emph{Deformable transformation} can map the shape of one image onto another through an elastic deformation model, allowing for nonlinear deformation in local regions to better adapt to different shape variations compared to rigid or affine transformation methods. In deformable transformation, we first define a deformation field that describes the amount of deformation at each pixel position and then solve for this deformation field to map one image onto another.

The model of deformable transformation can be simply represented as:
\begin{equation}
    \label{deformable_trans}
    \begin{bmatrix}
        x'\\y'
    \end{bmatrix}
    =
    \begin{bmatrix}
        x\\y
    \end{bmatrix}
    +
    \phi[x,~ y]
\end{equation}

Here, $\phi$ represents the deformation field, and $\phi[x, y]$ represents the transformation vector $(\Delta x,\Delta y)$ at $(x,y)$.

\subsection{Retinal Image Modalities}

To illustrate retinal image registration later, we introduce four commonly used techniques for photographing the eye: Color Fundus Photography (CF), Fluorescein Angiography (FA), Optical Coherence Tomography (OCT), and Optical Coherence Tomography Angiography (OCTA). These techniques provide various medical imaging tools to analyze retinal situations.

\subsubsection{Color Fundus Photography}
Color Fundus Photography (CF) involves using a fundus camera to capture color images of the retina using white light. Equipped with a low-power microscope, the camera magnifies the view of the eye's interior surface. This technique is cost-effective and straightforward for trained professionals \cite{BESENCZI2016371}. CF images (shown in Fig. \ref{examples}(a)) contain a broader range of fundus and rich color information, making it helpful in checking the atrophy of the retina and macular. Additionally, it helps diagnose retinopathy such as diabetes retinopathy, age-related macular degeneration, and glaucoma, as well as revealing signs of systemic diseases such as diabetes and cardiovascular diseases \cite{5660089}. 

\subsubsection{Fluorescein Angiography}
Fluorescein Angiography (FA), shown in Fig. \ref{examples}(b), involves a special dye called fluorescein and a camera to trace blood flow in the retina and choroid. It used a special dye, i.e., fluorescein, and a camera to examine blood flow in the retina and choroid. The radio-opaque dye is injected in a vein of the tester's arm while the retina vessels are photographed by tracing the dye before and after the injection. FA can detect capillary leakage \cite{fa1}, aneurysm, and neovascularization. However, some people may experience discomfort after the procedure \cite{FAintro}.

\subsubsection{Optical Coherence Tomography}
Optical Coherence Tomography (OCT) is an imaging technology that uses interference between an investigated object and a local reference signal to create high-resolution cross-sectional images and 3D scans of the retina and anterior segment \cite{oct}. Fig. \ref{examples}(c) shows a cross-sectional scan of OCT. It is a non-invasive technique that enables visualization of each layer of the retina, measurement of its thickness, and provides treatment guidance for conditions such as glaucoma, Diabetic Retinopathy (DR), and Age-related Macular Degeneration (AMD). Intra-operative OCT (iOCT) is necessary for many retinal therapies, including glaucoma surgery \cite{ioct1} and epiretinal device implantation \cite{ioct2}, as it provides real-time visualization of retinal layers.

\subsubsection{Optical Coherence Tomography Angiography}
\begin{figure*}
    \centering
    \subfigure[CF]{
        \begin{minipage}[b]{0.19\linewidth}
            \includegraphics[height=2.5cm]{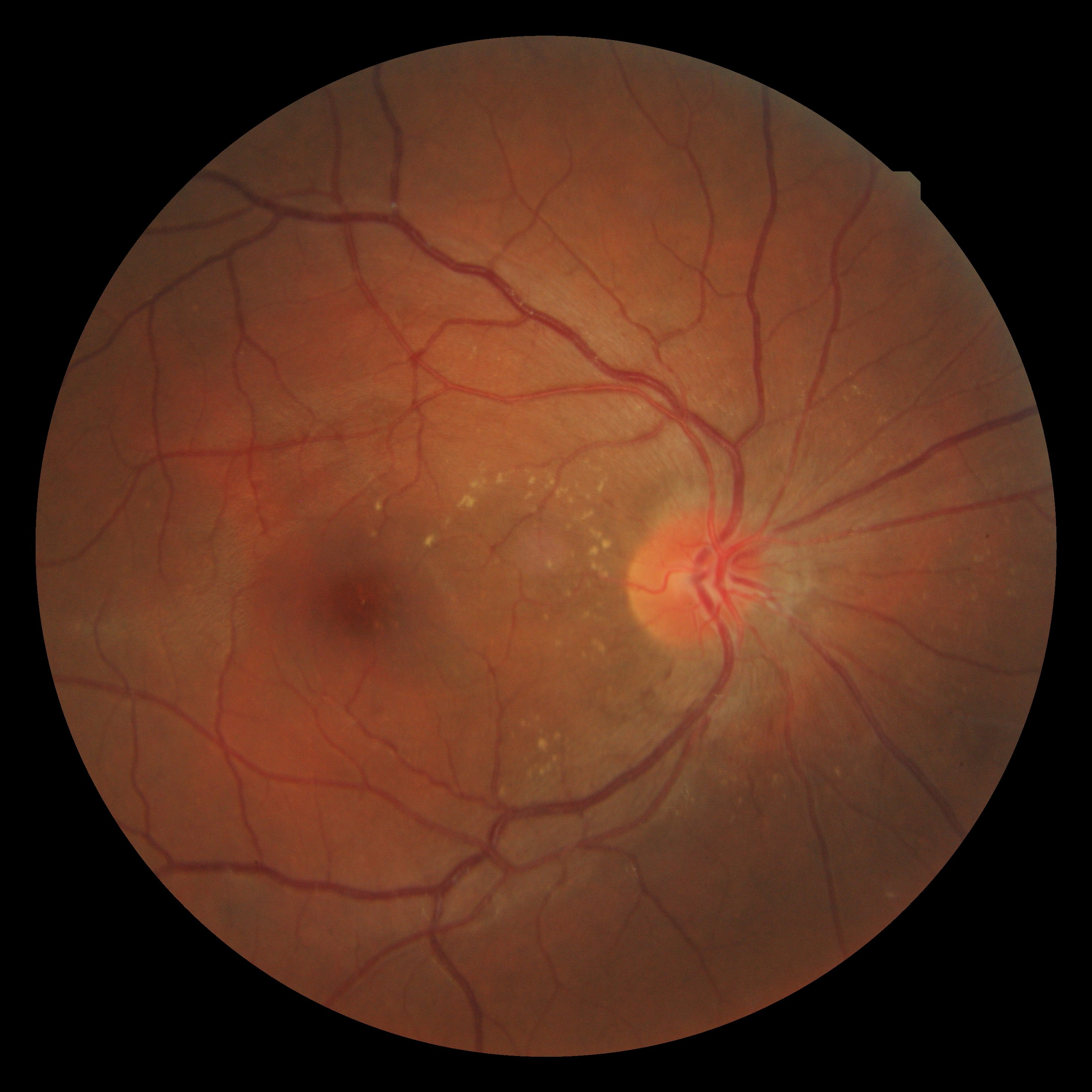}
        \end{minipage}
    }
    \subfigure[FA]{
        \begin{minipage}[b]{0.23\linewidth}
            \includegraphics[height=2.5cm]{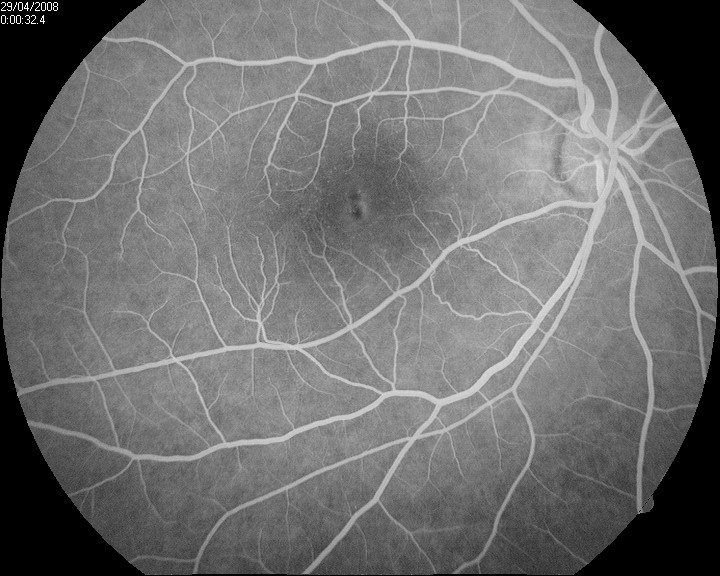}
        \end{minipage}
    }
    \subfigure[OCT]{
        \begin{minipage}[b]{0.28\linewidth}
            \includegraphics[height=2.5cm]{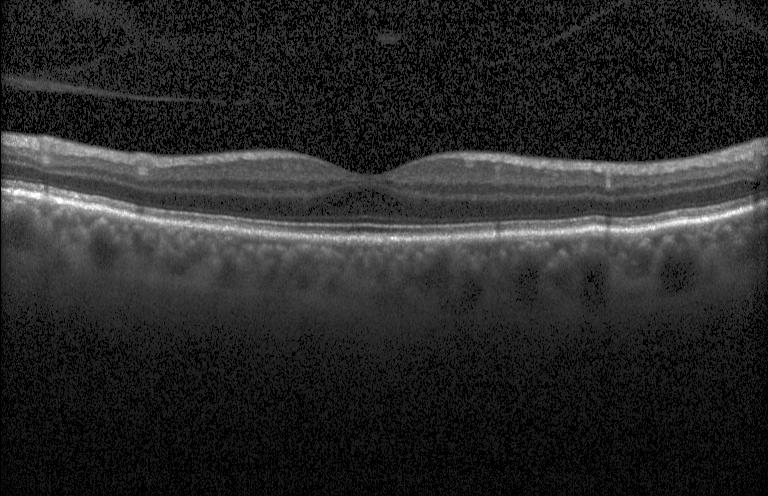}
        \end{minipage}
    }
    \subfigure[OCTA]{
        \begin{minipage}[b]{0.18\linewidth}
            \includegraphics[height=2.5cm]{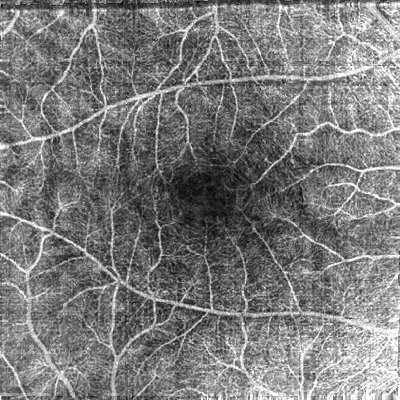}
        \end{minipage}
    }
    \caption{Fundus photography examples using different imaging techniques. (a) CF from FIRE dataset \cite{FIRE}. (b) FA from CF-FA dataset \cite{CF-FA}. (c) OCT from \cite{OCTdataset}. (d) OCTA from OCTA-500 dataset \cite{OCTA-500}. }
    \label{examples}
\end{figure*}

Fig. \ref{examples}(d) showcases Optical Coherence Tomography Angiography (OCTA), an emerging imaging technology that builds upon OCT. OCTA captures images of the vascular network with higher resolution and a smaller view than FA, without invasiveness. Using the decorrelation signal produced by moving red blood cells, OCTA generates an image of the microvascular network. Recent studies have demonstrated the ability of OCTA to surpass the limitations of assessing blood flow in the optic nerve and help explain the vascular pathogenesis of glaucoma \cite{octa_glaucoma} and show impressive success in preclinical DR diagnosis \cite{octa_dr}.

\section{Traditional Image Registration Methods}
\label{sec:mlm}

Researchers developed increasingly sophisticated algorithms and resilient features during the initial image registration phases to attain precise registration. The paper employs the phrase "traditional methods" to differentiate between techniques utilized before the advent of deep learning and those implemented after that.

\subsection{Intensity-based Methods}
\label{ibm}
\begin{figure}[h]
    \centering
    \includegraphics[scale=0.18]{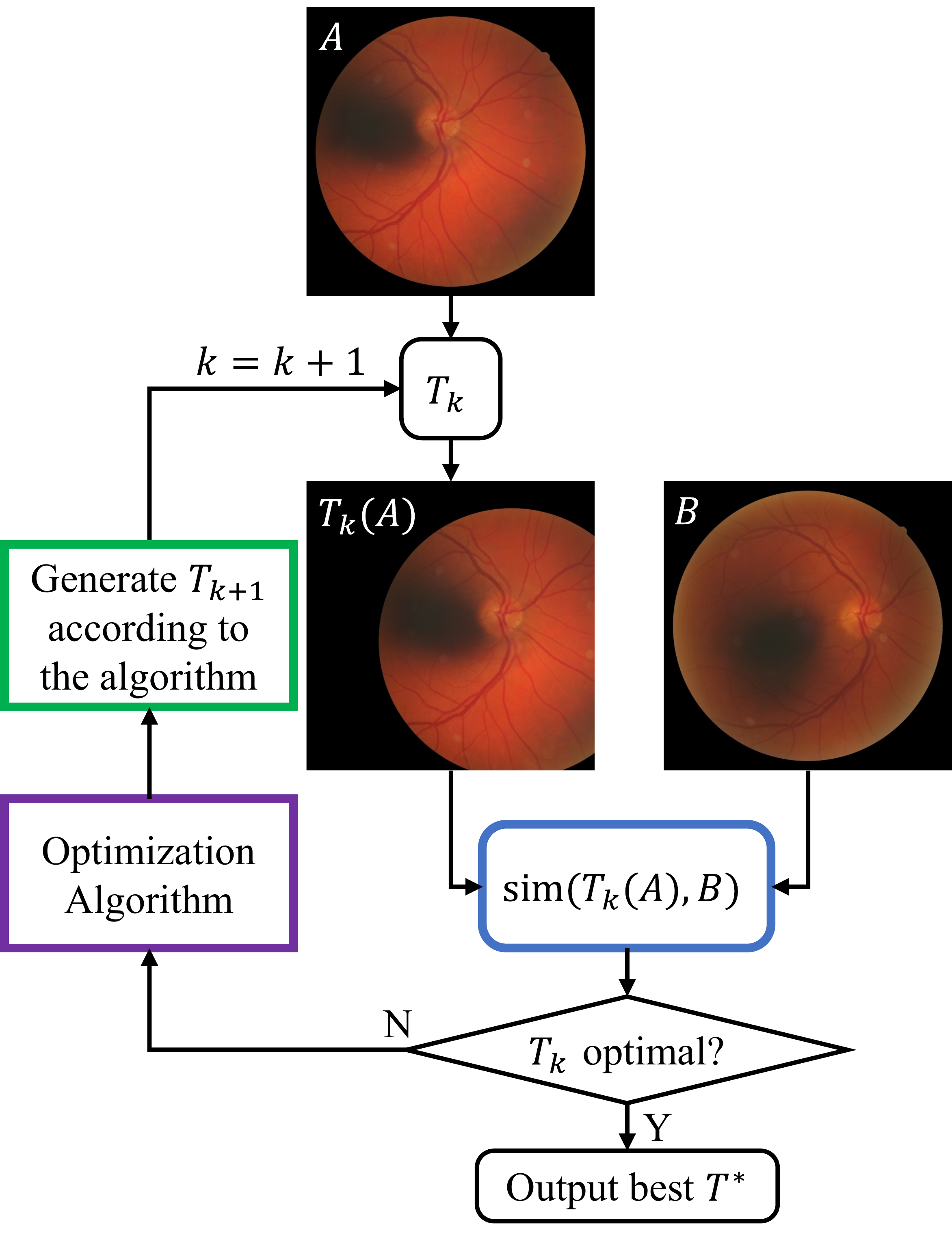}
    \caption{A general procedure of registration using iterative optimization}
    \label{ior}
\end{figure}

Intensity-based methods treat the problem as an iterative optimization problem. The basic steps of intensity-based registration are shown in Fig.\ref{ior}. Initially, a random transformation $T_0$ is selected, and an objective function is defined to measure the similarity between the transformed image $T_k(A)$ and the other image $B$. The goal is to find the optimal transformation $T^*$ to maximize the similarity. At each step, the optimization algorithm applies a perturbation to the parameters in $T$ based on the current similarity measure $\text{sim}(T_k(A), T(b))$. The process will be terminated when the similarity meets the requirement or converges with no more increase.

Researchers mainly concentrate on developing various similarity functions, including (normalized) cross-correlation (CC), (normalized) mutual information (MI), and sum of squared differences (SSD). These functions are typically calculated using the difference of each corresponding pixel of the input image pair. Among them, MI is considered the most important and widely used function. The LDDMM (Large Deformation Diffeomorphic Metric Mapping) \cite{beg2005computing} model is based on manifold learning theory and uses the Euler-Lagrange equation for optimization. It regards the image as a point on the manifold and achieves image registration by calculating the deformation between manifolds. This model can handle large deformations and maintain the nonlinear structure of the image.

Recently, there have been a few researches on intensity-based methods. Annkristin \emph{et al.} \cite{ib-1} proposed a normalized gradient fields (NGF) distance measure to deal with 2D-3D image registration. To overcome the drawback that using standard similarity measures may lead to optimization problems with many local optima, Öfverstedt \emph{et al.} \cite{ib-3} adopted a symmetric, intensity interpolation-free similarity measure combining intensity and spatial information. Castillo \cite{ib-2} proposed an intensity-based deformable image registration optimization formulation, making it easier to optimize. The similarity function is designed as a simple quadratic function formulation to be solved by straightforward coordinate descent iteration.

\subsection{Feature-based Methods}
\begin{figure}[h]
    \centering
    \includegraphics[scale=0.16]{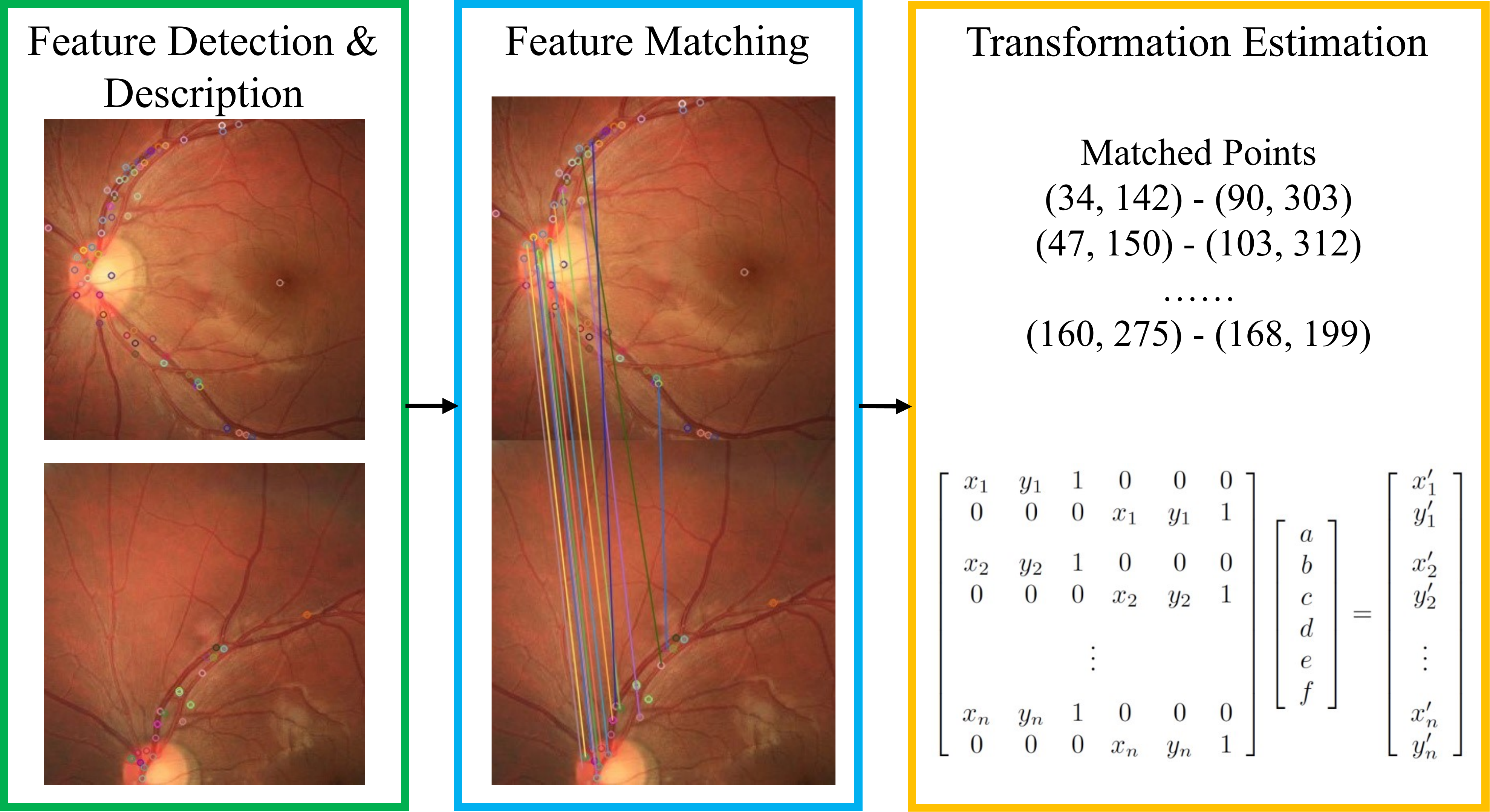}
    \caption{A general procedure of keypoint-based registration}
    \label{fbr}
\end{figure}

Feature-based methods are a popular way to match images based on their correspondence. These methods focus on local structures and salient features of images rather than global information. The process is divided into three steps. First, features such as points, edges, and regions are extracted from the input images. Next, a descriptor is calculated for each feature. In the matching stage, the closest features from the two images are matched to establish potential correspondences. The idea is that the corresponding points should have very similar descriptors. Finally, the transformation parameters are estimated based on the matching results. The primary challenge is determining the most effective method for extracting and describing features. Fig. \ref{fbr} illustrates the key point-based registration process.

One pioneering work in feature point-based registration is the Scale-Invariant Feature Transform (SIFT) \cite{sift}. SIFT transforms the image data into scale-invariant coordinates, identifies stable key points, assigns orientations to the key points, and generates feature descriptors for each key point. The extracted feature offers invariance under variations in scale, brightness, and angles. However, this process is often computationally expensive. To address this issue, various efforts \cite{surf, pcasift, daisy, orb, perspectivesift} have been made to enhance the performance and efficiency of SIFT. For instance, the Speeded Up Robust Features (SURF) \cite{surf} simplifies the filter function to reduce the dimension of descriptors and improve computational efficiency. Another method, the Oriented FAST and Rotated BRIEF (ORB) \cite{orb}, integrates the FAST \cite{fast} keypoint detector and BRIEF \cite{brief} descriptor to solve the high computational cost of SIFT features and the lack of rotation invariance, scale invariance and sensitivity to noise of the BRIEF feature. As a result, ORB is capable of delivering a speedup of up to two significant figures than SIFT. Other works focus on edge and contour features, using classic edge detection \cite{canny, log} and image segmentation \cite{pal1993review} algorithms for feature extraction.

\subsection{Retinal Applications}
In retinal image registration, intensity-based methods are first explored. The intensity similarity metrics mentioned above, such as MI \cite{legg2013improving, reel2013robust, reel2014enhanced} and CC \cite{340749}, are used.

The feature-based methods are more effective for retinal image registration than intensity-based methods. A popular approach is to utilize typical landmarks in retinal images. In 2003, Stewart \emph{et al.} \cite{db-icp} introduced the Dual-Bootstrap Iterative Closest Point (Dual-Bootstrap ICP) algorithm for retinal image registration. This algorithm starts by matching individual vascular landmarks and aligning images based on detected blood vessel centerlines. Other studies have also utilized vascular features \cite{guo2006tree, zheng2011landmark, ZHENG2014903, HERVELLA201897} and the optic disc \cite{koukounis2011retinal} for registration purposes.

One potential solution is to enhance the capabilities of key point detectors and feature descriptors to improve performance. Ramli \emph{et al.} \cite{ramli2017feature} designed a D-Saddle detector capable of detecting feature points even in low-quality regions. Yang \emph{et al.} \cite{4302762} built upon previous work \cite{db-icp} to create the generalized dual-bootstrap iterative closest point (GDB-ICP), which uses better initialization, robust estimation, and strict decision criteria to align retinal images from different modalities. Chen \emph{et al.} \cite{piifd} implemented a Harris detector to identify corner points, extract partial intensity invariant feature descriptors (PIIFD), and perform bilateral matching between image pairs. Outliers are then removed, followed by applying the final transformation. Ramli \emph{et al.} \cite{ramli2017feature} improved the Saddle detector to detect feature points for low-quality regions. Gharabaghi \emph{et al.} \cite{gharabaghi2013retinal} utilized affine moment invariants (AMI) as a shape descriptor. Combining domain knowledge, SIFT and its variants are used in \cite{ghassabi2015colour, saha2016twostep}. Li \emph{et al.} \cite{li2020orientation} introduced Orientation-independent Feature Matching (OIFM) that uses a new circular neighborhood-based feature descriptor.

\subsection{Analysis}
In the traditional registration stage, there are many applications for retinal image registration, and many registration methods directly use various retinal modalities as evaluation indicators. Some work also specifically introduces domain knowledge based on some general methods in retinal image registration. The intensity-based approach can be sensitive to the intensity distribution when the image pair has varying illumination due to different cameras, modalities, or retinopathy-induced background changes. Feature-based methods also suffer from this problem because the feature needs descriptors. Another drawback is that most traditional registration methods take much longer for inference.

\section{Deep Learning-based Registration Methods}
\label{sec:dlm}
Deep learning-based image segmentation has proved to be a robust tool in image segmentation since 2019 \cite{hesamian2019deep}. These methods can improve accuracy and efficiency by automatically learning high-level features from input images. The registration task, similar to the segmentation, has thus been developed utilizing deep learning methods. 
They differ from feature-based approaches by utilizing deep neural networks to replace the feature extractor, feature matching process, and transformation process. Rather than directly optimizing transformation parameters, these methods indirectly optimize registration model parameters, revealing the true essence of their effectiveness.

\subsection{Feature-based Methods}
The convolutional network (CNN) is a pioneering work in computer vision. It uses learnable convolution kernels and inductive bias, such as locality and translation equivariance, to detect learned patterns in local regions and extract high-level features. This characteristic makes CNNs especially suitable for object detection and image registration tasks, where spatial features are essential. Table \ref{cnn_table} displays prominent works in CNN-based registration methods, which have become the most popular approach in the field since 2016.

\begin{table*}
    \caption{Overview of feature-based image registration methods. For the supervision column, S is for supervised, W is for weakly supervised, and U is for unsupervised. For the last column, MM is for multi-modal.}
    \centering
    \resizebox{\linewidth}{!}{
    \begin{tabular}{llllllllll}
        \toprule
        \textbf{Method} & \textbf{Year} & \textbf{Scene} & \textbf{Dimension} & \textbf{Modality} & \textbf{Type} & \textbf{Supervision} & \textbf{MM}\\
        \midrule
        Miao \emph{et al.} \cite{miao2016real} & 2016 & Virtual & 2D/3D & X-ray/CT & Rigid & S & N\\
        Quicksilver \cite{quicksilver} & 2017 & Brain & 3D & MR & Deformable & S & N\\
        Cao \emph{et al.} \cite{cao2017deformable} & 2017 & Brain & 3D & MR & Deformable & S & N\\
        DIRNet \cite{DIRNet} & 2017 & Digits/Heart & 2D & Handwritten/MR & Deformable & U & N\\
        Li \emph{et al.} \cite{li2017non} & 2017 & Brain & 3D & MR & Deformable & U & N\\
        Zheng \emph{et al.} \cite{zheng2018pairwise} & 2018 & Bone & 2D/3D & X-ray/CT & Rigid & S & Y\\
        Sloan \emph{et al.} \cite{sloan2018learning} & 2018 & Brain & 3D & MR & Rigid & S & N\\
        AIRNet \cite{chee2018airnet} & 2018 & Brain & 3D & MR & Affine & S & N\\
        Lv \emph{et al.} \cite{lv2018respiratory} & 2018 & Abdominal & 3D & MR & Deformable & S & N\\
        Hu \emph{et al.} \cite{HU20181} & 2018 & Prostate gland & 3D & MR/US & Deformable & W & Y\\
        
        Jiang \emph{et al.} \cite{jiang2018cnn} & 2018 & Chest & 3D & CT & Deformable & U & N\\
        Li \emph{et al.} \cite{li2018non} & 2018 & Brain & 3D & MR & Deformable & U & N\\
        BIRNet \cite{FAN2019193} & 2019 & Brain & 3D & MR & Deformable & S & N\\
        DeepAtlas \cite{xu2019deepatlas} & 2019 & Knee/Brain & 3D & MR & Deformable & W & N\\
        DLIR \cite{de2019deep} & 2019 & Heart/Chest & 3D & MR/CT & Affine/Deformable & U & N\\
        VTN \cite{zhao2020vtn} & 2019 & Liver/Brain & 3D & CT/MR & Affine/Deformable & U & N\\
        Zhao \emph{et al.} \cite{Zhao_2019_ICCV} & 2019 & Liver/Brain & 3D & CT/MR & Affine/Deformable & U & N\\
        VoxelMorph \cite{voxelmorph} & 2019 & Brain & 3D & MR & Deformable & W/U & N\\
        VoxelMorph-diff \cite{dalca2019unsupervised} & 2019 & Brain & 3D & MR & Deformable & U & N\\
        Dual-PRNet \cite{hu2019dual} & 2019 & Brain & 3D & MR & Deformable & U & N\\
        
        DeepFLASH \cite{Wang_2020_CVPR} & 2020 & Synthetic/Brain & 2D/3D & Eye/MR & Deformable & S & N\\
        Mansilla \emph{et al.} \cite{mansilla2020learning} & 2020 & Chest & 2D & X-ray & Deformable & W & N\\
        Mok \emph{et al.} \cite{mok2020fast} & 2020 & Brain & 3D & MR & Deformable & U & N\\
        CycleMorph \cite{kim2021cyclemorph} & 2021 & Face/Brain & 2D/3D & Expression/MR & Deformable & U & N\\
        DeepSim \cite{czolbe2021deepsim} & 2021 & Brain/Cell & 3D & MR/EM & Deformable & W/U & N\\
        Mok \emph{et al.} \cite{mok2022robust} & 2022 & Brain & 3D & MR & Deformable & U & N\\
        Dual-PRNet$^{++}$ \cite{kang2022dual} & 2022 & Brain & 3D & MR & Deformable & U & N\\
        Tran \emph{et al.} \cite{tran2022light} & 2022 & Liver/Brain & 3D & CT/MRI & Deformable & U & N\\
        IMSE \cite{kong2023indescribable} & 2023 & Brain & 2D/3D & CT/MR & Deformable & U & Y\\
        Zhang \emph{et al.} \cite{zhang2023alternately} & 2023 & Liver & 3D & US & Deformable & U & N\\ %
        
        AMNet \cite{che2023amnet} & 2023 & Brain & 3D & MR & Deformable & U & N\\ %
        
        \midrule
        DeepSPa \cite{Lee_2019_ICCV} & 2019 & \multirow{15}{1.5cm}{Retina} & \multirow{15}{1.5cm}{2D} & CF / FA / OCT & Affine & U & Y\\
        Zhang \emph{et al.} \cite{zhang2019joint} & 2019 & & & CF / FA & Deformable & W & Y\\
        Silva \emph{et al.} \cite{silva2020deep} & 2020 & & & CF / FA / IR & Affine & S & Y\\
        Wang \emph{et al.} \cite{wang2020segmentation} & 2020 & & & CF / IR & Perspective & S & Y\\
        Tian \emph{et al.} \cite{tian2020multi} & 2020 & & & CF / OCT & Deformable & U & Y\\
        Zou \emph{et al.} \cite{zou2020non} & 2020 & & & CF & Deformable & U & N\\
        Wang \emph{et al.} \cite{wang2021robust} & 2021 & & & CF / FA / IR & Perspective & S & Y\\
        Zhang \emph{et al.} \cite{zhang2021two} & 2021 & & & CF / FA / IR & Affine/Deformable & S & Y\\
        Sui \emph{et al.} \cite{sui2021deep} & 2021 & & & MSI & Deformable & W & N\\
        An \emph{et al.} \cite{an2022self} & 2022 & & & CF / FA / IR & Rigid & U & Y\\
        
        Benvenuto \emph{et al.} \cite{benvenuto2022unsupervised} & 2022 & & & CF & Deformable & U & N\\
        Lopez \emph{et al.} \cite{lopez2022unsupervised} & 2022 & & & OCTA & Deformable & U & N\\
        Rivas \emph{et al.} \cite{rivas2022color} & 2022 & & & CF & Similarity & S & N\\
        Kim \emph{et al.} \cite{kim2022robust} & 2022 & & & CF & Perspective & S & N\\
        Rivas \emph{et al.} \cite{rivas2023automated} & 2023 & & & OCT & Affine + Z-axis & U & N\\
        
        \bottomrule
    \end{tabular}
    }
    \label{cnn_table}
\end{table*}

\begin{figure}[h]
    \centering
    \includegraphics[scale=0.2]{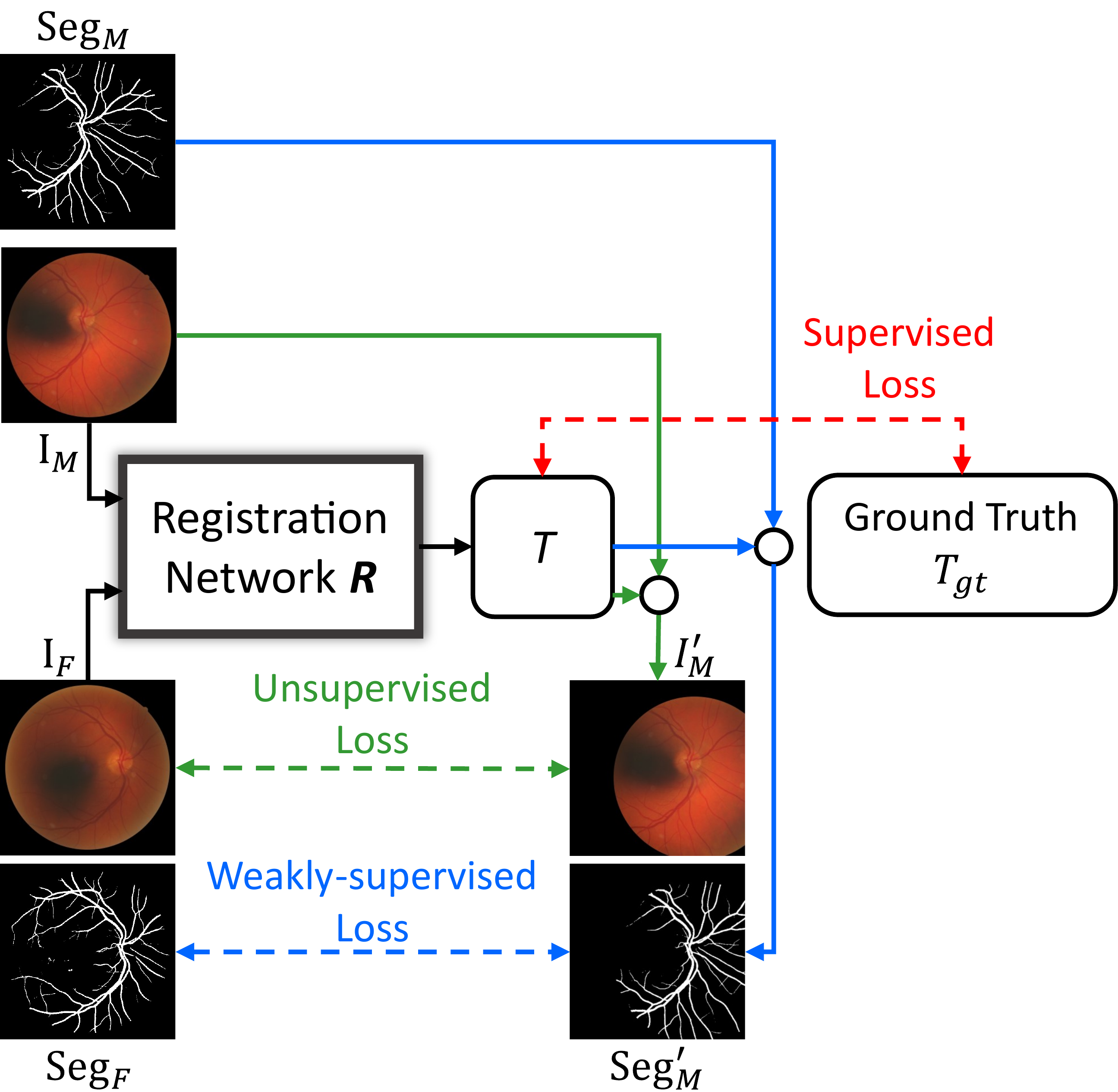}
    \caption{The overall framework for end-to-end deep learning-based medical image registration methods. Red, blue, and green lines denote the supervised, weakly-supervised, and unsupervised training strategies, respectively. The small circles denote performing spatial transformation with the predicted transform $T$ using STN \cite{STN}.}
    \label{cnnreg}
\end{figure}

\subsubsection{Patch-based Methods}
Instead of direct regression of registration parameters from the image pair, the patch-based approach was used, dividing the image into smaller patches. The patch is utilized in different ways depending on the predicted transformation type. In the case of linear transformations, the network establishes a match that can be used to derive the registration parameters. Conversely, a local displacement field is outputted and combined for nonlinear transformations. Various CNN models were proposed by Zagoruyko \emph{et al.} \cite{zagoruyko2015learning} that output the similarity of two image patches as feature descriptors. Cao \emph{et al.} \cite{cao2017deformable} proposed a similarity-steered CNN regression architecture that estimates the displacement vectors at each corresponding location between linearly aligned brain MR image pairs. An interpolation is subsequently utilized to obtain the dense deformation. Lv \emph{et al.} \cite{lv2018respiratory} divided the signal into three bins and used CNN to estimate the displacement field for abdominal motion correction throughout the respiratory cycle. However, these methods typically require an additional step of patch selection and final registration, which can be time-consuming. Additionally, generating or manually labeling ground truth can be a limiting factor.

\subsubsection{End-to-end CNN Methods}
Due to increased computing power, supervised end-to-end networks are developed for direct registration. The ground truth is obtained by traditional algorithms or manual labels. A general end-to-end deep learning registration method framework is shown in Figure \ref{cnnreg}. Miao \emph{et al.} \cite{miao2016real} employed 2D/3D CNN regressors to estimate the rigid transformation parameters in real time directly. Quicksilver \cite{quicksilver} divides the 3D brain MRI into 3D patches due to the limitation of GPU memory, but it can directly predict the deformation field for the input patches. To improve the performance of supervised methods, Chee \emph{et al.} \cite{chee2018airnet} leveraged unlabelled data to generate a synthetic dataset and further trained the AIRNet (affine image registration network) based on it. BIRNet \cite{FAN2019193} was proposed as a hierarchical dual-supervised fully convolutional neural network based on U-Net \cite{UNet} in the following year, with a loss function designed as a combination of the difference in image intensity and the difference of predicted displacement and ground truth displacement in each layer of U-Net's decoder. Wang \emph{et al.} \cite{Wang_2020_CVPR} introduced a low dimensional Fourier representation of diffeomorphic transformations to improve training and inference efficiency.

Weakly supervised registration methods take advantage of additional semantic information to ensure meaningful registration, and they also overcome the challenge of the unavailability of ground truth transformation. These methods utilize extra information, such as anatomical segmentation, to perform registration. Hu \emph{et al.} \cite{HU20181} proposed a weakly supervised registration network for multi-modal 3D prostate gland images using the ground truth segmentation label of the gland and other anatomical landmarks. Xu \emph{et al.} \cite{xu2019deepatlas} proposed a deep learning framework named DeepAtlas that jointly learns networks for image registration and segmentation, which are trained alternately, complementing each other to achieve better results with only a few labels for segmentation.

Unsupervised methods have also been researched to eliminate any ground truth labels further. Spatial transformer layer (STL) \cite{STN}, a differentiable module that can warp the input image, has been the foundation of many unsupervised registration methods. STL enables obtaining the transformed moving image in a differentiable manner, which allows applying conventional similarity measurement between the transformed and fixed images during training as the loss function. In 2017, DIRNet \cite{DIRNet} was introduced as the first end-to-end unsupervised deformable registration network by adopting STL. Later, VoxelMorph \cite{voxelmorph} was proposed as a U-Net-based network that achieved faster run time and better performance than traditional iterative-based methods, with only unsupervised training. Auxiliary anatomical segmentation can be optionally used in a weakly-supervised setting. In their following work, Dalca \emph{et al.} \cite{dalca2019unsupervised} further adopted a probabilistic generative model to provide diffeomorphic guarantees. Dual-PRNet \cite{hu2019dual} extended VoxelMorph \cite{voxelmorph} by incorporating a pyramid registration module that uses multi-level context information and sequentially warps the convolutional features. Dual-PRNet$^{++}$ \cite{kang2022dual} further enhanced the PR module in Dual-PRNet by computing correlation features and using residual convolutions.

\subsubsection{Deep Similarity Methods}
In deep learning, pixel-based similarity metrics like MSE and NCC are commonly employed. However, these metrics may sometimes encounter difficulties when dealing with low-intensity contrast or noise. To address these issues, deep similarity methods have been developed, which utilize a custom similarity measure. For example, DeepSim \cite{czolbe2021deepsim} utilizes semantic information extracted by a pre-trained feature extractor in a segmentation network to construct a semantic similarity metric. This specialized metric allows the network to learn and adapt to dataset-specific features, improving low-quality image performance. IMSE \cite{kong2023indescribable} takes it a step further with a self-supervised approach to train a modality-independent evaluator using a new data augmentation technique called shuffle remap, which can provide style enhancement. The evaluator then serves as a multi-modal similarity estimator to train a multi-modal registration network.

\subsubsection{Cascade Methods}
Cascade methods are inspired by traditional iterative registration. The cascade architecture, namely, stacking networks in series, can provide progressive registration in a coarse-to-fine manner. DLIR \cite{de2019deep} implemented a cascade architecture by stacking an affine network followed by multiple deformable networks, with each network being trained sequentially with the weights of previous networks fixed. In contrast, Zhao \emph{et al.} \cite{zhao2020vtn, Zhao_2019_ICCV} proposed a recursive cascade architecture similar to DLIR but much more sophisticated. They jointly trained their cascade networks to learn progressive alignments more effectively.

\subsubsection{Consistency-based Methods}
Consistency-based methods add consistency constraints based on the property of registration or transformation. In 2020, Mok \emph{et al.} tackled the challenge of the deformable transformation's invertibility by introducing a swift and symmetrical diffeomorphic image registration approach \cite{mok2020fast}. The network was trained with an inverse-consistency constraint, enabling it to learn bidirectional transformations to the mean shape of two input images to produce topology-preserving and inverse-consistent transformations. The following year, Kim \emph{et al.} proposed CycleMorph \cite{kim2021cyclemorph}, which utilized cycle consistency as an additional constraint to enhance topology preservation and reduce folding issues. To register images X to Y and Y to X, the method employs two CNNs, $G_X$ and $G_Y$. The warped images from both networks are used as image pairs and sent to the networks themselves to ensure they can be returned to their original state, maximizing the similarity between the original image and the reversed image.

\subsubsection{Other Methods}
However, with the development of novel architectures, the parameters increased significantly, so it is harder to achieve real-time registration without high computing power. Tran \emph{et al.} \cite{tran2022light} tried to solve this problem via knowledge distillation. They transferred meaningful knowledge of distilled deformations from a pre-trained high-performance network (teacher network) to a fast, lightweight network (student network). After training, only the lightweight student network is used during the inference, allowing the model to achieve fast inference time using only a common CPU.

\subsubsection{Retinal Applications}
Utilizing retinal landmarks inspired researchers to develop new techniques for detecting these landmarks automatically. In particular, \cite{Lee_2019_ICCV} used handcrafted features, while \cite{rivas2022color, kim2022robust} employed CNNs. Lee \emph{et al.} employed a CNN to classify patches of various step patterns with intensity changes. On the other hand, Rivas \emph{et al.} \cite{rivas2022color} used a CNN to produce a heatmap of blood vessels and bifurcations and applied the maxima detection and feature matching method RANSAC \cite{fischler1981ransac} during testing. Similarly, Kim \emph{et al.} \cite{kim2022robust} used a vessel segmentation network and a joint detection network to identify vascular landmark points for registration. The SIFT algorithm \cite{sift} was then used to compute descriptors based on the region around these points. Benvenuto \emph{et al.} \cite{benvenuto2022unsupervised} used Isotropic Undecimated Wavelet Transform, which segments blood vessels and ocular shape. Based on the segmentation, the registration network adopted from U-Net is trained to perform registration. This year, Rivas \emph{et al.} \cite{rivas2023automated} also explored deep learning registration methods for OCT 3D Scan. They first performed affine alignment on a 2D projection and then z-axis registration based on layer segmentation.

Recent studies have explored the potential of end-to-end methods utilizing innovative network architectures. Silva \emph{et al.} \cite{silva2020deep} utilized a VGG 16 feature extractor, correlation matrix, and regression network to replicate the traditional steps of feature-based registration, including feature extraction, matching, and registration transform. They evaluated the model's effectiveness on a multi-modal retinal dataset. Meanwhile, Tian \emph{et al.} \cite{tian2020multi} improved U-Net \cite{UNet} using image pyramid as multi-scale input and introduced a new edge similarity loss calculated via the correlation between gradients of the fixed and moved image. Still based on U-Net, Sui \emph{et al.} \cite{sui2021deep} further sent an image pyramid of the original image and the ground truth vessel map into each layer of encoder and decoder, respectively. 

It is worth noting that Wang \emph{et al.} \cite{zhang2019joint, wang2020segmentation, wang2021robust, zhang2021two, an2022self} made significant strides in multi-modal retinal image registration. They began with a deformable registration model, which included a vessel segmentation network and a deformation field estimation network, in \cite{zhang2019joint}. In their later work \cite{wang2020segmentation}, they adapted the vessel segmentation network from \cite{zhang2019joint} and further used a pre-trained SuperPoint \cite{detone2018superpoint} model for feature detection and description, along with an outlier rejection network for perspective registration. They utilized this three-stage methodology (segmentation, detection and description, and outlier rejection) in their subsequent research. The advantage of this approach is that it bypasses the intensity gap of different modalities, but the downside is its high complexity. They improved the segmentation network with pixel-adaptive convolution in \cite{wang2021robust}. In \cite{zhang2021two}, they utilized perspective registration as coarse registration and added a deformable framework for fine alignment, achieving remarkable accuracy. Finally, their latest work \cite{an2022self} made the three-stage methodology a self-supervised one.

\subsubsection{Analysis}
Limited by insufficient computing power in the early stage, patch-based methods appeared earlier. As computing power and network diversity increase, it has been able to take the entire image and even 3D images as input and perform feature extraction and matching tasks integrated and simultaneously. During this period, thanks to the rapid advances in deep learning architecture, the feature extraction module for image registration has also been steadily developing along with the trend. In addition, many novel constraints have been proposed. These constraints exploit some fundamental properties of the registration or transformation, resulting in a more suitable shape for the output transformation.

Deep learning-based retinal image registration did not appear until 2019. There are two types of approaches to this method. The first type employs mainstream registration methods which utilize advanced network architectures. Although these methods achieve exceptional performance, they rely heavily on the design of network architecture rather than domain knowledge. The second type of approach focuses on solving the registration problem in a more domain-specific manner. These methods extract or leverage critical features such as vessel segmentation or vascular junctions for later registration. 

We have observed that the diversity of retinal image registration works is notably lower when compared to other medical registration works. This can be attributed to a few factors, including imaging principles and targets. Regarding imaging principles, CT and MR use X-rays and magnetic fields. These imaging techniques can produce high tissue contrast while maintaining the intensity consistency of different images. However, when it comes to retinal image registration, CF and FA images are commonly used, which rely solely on white light illumination for imaging. This imaging principle, coupled with the natural movement of the subject's eyeballs, can lead to differences in brightness when taking multiple shots. The second factor is the imaging target. CT and MR are typically used for imaging the chest, abdomen, and brain, with many features such as multiple organs, tissues, and brain regions. In contrast, retinal images mainly focus on the blood vessel tree and optic disc, with features that have relatively low discrimination. As a result, learning robust features in retinal image registration automatically proves to be a more challenging task.

\subsection{Translation-based Methods}

\begin{table*}
    \caption{Overview of Translation-based image registration methods}
    \centering
    \resizebox{\linewidth}{!}{
    \begin{tabular}{clllllll}
        \toprule
        \textbf{Methodology} & \textbf{Method} & \textbf{Year} & \textbf{Scene} & \textbf{Dimension} & \textbf{Modality} & \textbf{Type}\\
        \midrule
        \multirow{5}{1.5cm}{GAN}
        & Mahapatra \emph{et al.} \cite{mahapatra2018deformable} & 2018 & Retina/Heart & 2D & CF/FA/MR & Deformable\\
        & Qin \emph{et al.} \cite{qin2019unsupervised} & 2019 & Lung/Brain & 2D & CT/MR & Deformable\\
        & Xu \emph{et al.} \cite{xu2020adversarial} & 2020 & Kidney/Abdomen & 3D & CT/MR & Deformable\\
        & Han \emph{et al.} \cite{han2022deformable} & 2022 & Brain & 3D & MR/CT & Deformable\\
        & MedRegNet \cite{santarossa2022medregnet} & 2022 & Retina & 2D & CF/FAF/FAG & Perspective\\
        \midrule
        \multirow{2}{1.5cm}{Contrastive\\Learning}
        & Casamitjana \emph{et al.} \cite{Casamitjana_2021} & 2021 & Brain & 3D & Histology/MRI & Deformable\\
        & Chen \emph{et al.} \cite{chen2022unsupervised} & 2022 & Thorax/Abdomen/Lung & 3D & CT/MRI & Deformable\\
        \midrule
        DDPM & DiffuseMorph \cite{kim2022diffusemorph} & 2022 & Face/Brain & 2D/3D & Expression/MR & Deformable\\
        \bottomrule
    \end{tabular}
    }
    \label{translation_table}
\end{table*}

\begin{figure}[h]
    \centering
    \includegraphics[scale=0.16]{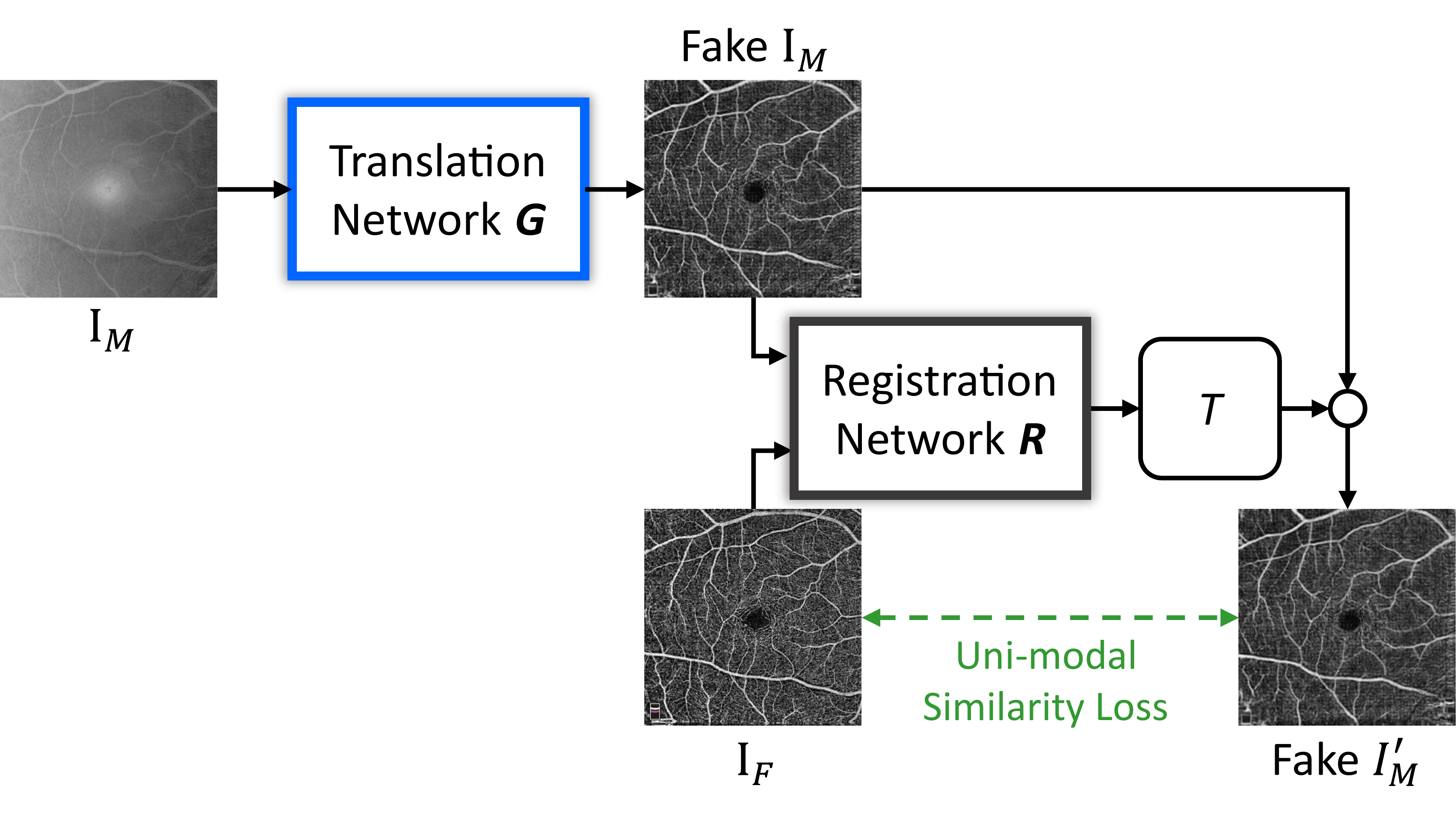}
    \caption{The overall framework for translation-based methods.}
    \label{transreg}
\end{figure}

Multi-modal image registration can be complex as it involves aligning images of varying modalities with unique intensity distributions. This can pose a challenge for uni-modal methods. However, an innovative solution to this issue is to leverage image translation techniques. This solution transforms the multi-modal registration problem into a more straightforward uni-modal registration problem, as depicted in Fig. \ref{transreg}. Table \ref{translation_table} highlights the top translation-based registration algorithms available.

\subsubsection{Generative Adversarial Network}

Generative adversarial network (GAN) \cite{NIPS2014gan} consists of two sub-networks, a generator and a discriminator, trained in a game-theoretic setting to generate synthetic data indistinguishable from the actual data. The generator generates synthetic samples while the discriminator attempts to differentiate between natural and synthetic samples, and the training process continues until the generated samples are indistinguishable from actual ones.

Mahapatra \emph{et al.} \cite{mahapatra2018deformable} used a GAN to generate the registered image with the identical distribution of the moving image and the deformation field. They also ensured that the structure of the generated image matched that of the reference image through structural similarity loss. Qin \emph{et al.} \cite{qin2019unsupervised} proposed a method of decomposing images into a latent shape space and a separate latent appearance space for both modalities, which was used to learn a bi-directional registration function.

CycleGAN \cite{Zhu_2017_ICCV}, which is based on GAN, enables image-to-image (i2i) translation using unpaired images. It employs a cycle consistency loss to ensure the reconstructed images are consistent with the original input images.
Several multi-modal registration methods \cite{xu2020adversarial, han2022deformable} used CycleGAN as the primary network for image translation. Xu \emph{et al.} \cite{xu2020adversarial} introduced two additional losses to enforce structural similarity between the translated and authentic images. They also jointly trained translated uni-modal and multi-modal streams to complement each other. Han \emph{et al.} \cite{han2022deformable} implemented image synthesis in both directions and predicted the associated uncertainty, providing information used in the fusion of the two direction estimations.

\subsubsection{Contrastive Learning}
Contrastive learning defines positive and negative samples, and the goal is to learn a representation space where positive samples are close to each other while negative ones are far away. A recent study by Park \emph{et al.} \cite{park2020contrastive} explored integrating contrastive learning into image translation by introducing an extra loss called PatchNCE to naive GAN. This loss encourages generated output patches to be closer to their corresponding image patches than random ones. Casamitjana \emph{et al.} \cite{Casamitjana_2021} used the PatchNCE loss to train an i2i translation network for transferring source images to the desired target domain. They then applied an independently trained intra-modality registration network in the target domain to predict the deformation field. Building on this work, Chen \emph{et al.} \cite{chen2022unsupervised} proposed an end-to-end architecture that jointly trained the registration and translation network without requiring a discriminator.

\subsubsection{Denoising Diffusion Probabilistic Model}
A new generative model called the denoising diffusion probabilistic model (DDPM) \cite{ho2020denoising} was recently introduced. This model is designed to learn the Markov transformation from a simple Gaussian distribution to the actual data distribution. DDPM has been shown to generate images of higher quality than GAN \cite{dhariwal2021diffusion}. Additionally, Kim \emph{et al.} \cite{kim2022diffusemorph} developed DiffuseMorph, the first and currently only registration network based on diffusion. The network estimates the score function by adding a diffusion network before a standard registration network, and it even shows the image registration trajectory by scaling the conditional score. However, unlike translation between modalities, DiffuseMorph constructs the score function directly between the input image pair.

\begin{table*}
    \caption{Overview of Transformer-based image registration methods}
    \centering
    \resizebox{\linewidth}{!}{
    \begin{tabular}{llllllll}
        \toprule
        \textbf{Method} & \textbf{Year} & \textbf{Scene} & \textbf{Dimension} & \textbf{Modality} & \textbf{Type} & \textbf{MM}\\
        \midrule
        ViT-V-Net \cite{chen2021vit} & 2021 & Brain & 3D & MR & Deformable & N\\
        Zhang \emph{et al.} \cite{zhang2021learning} & 2021 & Brain & 3D & MR & Deformable & N\\
        C2FViT \cite{mok2022affine} & 2022 & Brain & 3D & MR & Affine & N\\
        TransMorph \cite{chen2022transmorph} & 2022 & Brain/Heart & 3D & MRI/XCAT/CT & Affine/Deformable & N\\
        TD-Net \cite{song2022td} & 2022 & Brain & 3D & MR & Deformable & N\\
        Wang \emph{et al.} \cite{wang2022transformer} & 2022 & Brain & 3D & MR & Deformable & N\\
        XMorpher \cite{shi2022xmorpher} & 2022 & Heart & 3D & CT & Deformable & N\\
        Swin-VoxelMorph \cite{zhu2022swin} & 2022 & Brain & 3D & MR & Deformable & N\\
        TransMatch \cite{chen2023transmatch} & 2023 & Brain & 3D & MR & Deformable & N\\
        ModeT \cite{wang2023modet} & 2023 & Brain & 3D & MR & Deformable & N\\
        \bottomrule
    \end{tabular}
    }
    \label{transformer_table}
\end{table*}

\subsubsection{Retinal Applications}
Although there have been many studies on image-to-image translation in various retinal modalities \cite{kamran2020fundus2angio, andreini2021two}, few studies on retinal image registration use translation-based techniques. While MedRegNet \cite{santarossa2022medregnet} is the only method that utilizes CycleGAN \cite{Zhu_2017_ICCV} as an image translation tool, it is solely used as a multi-modal retinal data generator. On the other hand, \cite{zhang2019joint, wang2020segmentation, wang2021robust, zhang2021two, an2022self} mentioned in the previous part can also be considered as translation-based methods when dealing with multi-modal data. They employ image segmentation to obtain blood vessel segmentation maps, which can be viewed as converting different modalities into a single "mask" modality.

\subsubsection{Analysis}
Translation-based methods have been shown to effectively solve multi-modal problems by transforming image pairs into the same modality, thus reducing registration difficulty. Recently, registration methods using different types of generative networks have emerged with the updated generative network models. While GAN can produce good results in modality translation, its training process is quite complex, requiring more time-consuming manual hyperparameter adjustment for the generator and discriminator. In the past, contrastive learning accounted for half of the unsupervised field, but it typically requires large amounts of high-quality data for training. Diffusion is another recently emerging image generation method that can produce quite realistic effects, but its applicability in registration still needs to be explored. Unfortunately, research progress in this area is limited, and the need for more public datasets for retinal image registration may be one of the reasons.

\subsection{Transformer-based Methods}
Recently, Google explored a way to use pure transformer architecture in vision tasks, known as Vision Transformer (ViT) \cite{dosovitskiy2021an}, outperforming existing CNN methods' performance. ViTs split the image into patches and treat them like tokens as in an NLP application, which has led to its successful application in various computer vision tasks, including image registration. Table \ref{transformer_table} overviews transformer-based image registration methods.

\subsubsection{Hybrid Methods}
In the beginning, researchers attempted to integrate Transformers into CNN-based models. Chen \emph{et al.} \cite{chen2021vit} pioneered the use of ViT on high-level features extracted from convolutional layers of moving and fixed images. Building on this approach, Song \cite{song2022td} proposed TD-Net, which utilizes multiple Transformer blocks for down-sampling. Conversely, Zhang \emph{et al.} \cite{zhang2021learning} introduced a dual Transformer network comprised of two branches - intra-image and inter-image - with Transformers embedded in both branches to enhance features, similar to the approach in \cite{chen2021vit}. Wang \emph{et al.} \cite{wang2022transformer} enhanced the UNet \cite{UNet} architecture for registration by introducing a bi-level connection and a unique Transformer block. TransMorph \cite{chen2022transmorph} was then proposed as a hybrid Transformer-ConvNet model, utilizing Swin Transformers \cite{liu2021swin} in the encoder and convolutional layers in the decoder. The authors demonstrated that positional embedding could be disregarded, leading to a flatter loss landscape for registration. The following year, Chen \emph{et al.} \cite{chen2023transmatch} proposed TransMatch, emphasizing the importance of inter-image feature matching. They employed a Transformer-based encoder and matched regions using their new local window cross-attention module. Recently, Wang \emph{et al.} \cite{wang2023modet} introduced a motion decomposition transformer (ModeT) based on a multi-head neighborhood attention mechanism which can model multiple motion modalities.

\subsubsection{Complete Transformer Methods}
An alternative method involves integrating a complete Transformer architecture into the network. In a recent study by Shi \emph{et al.} \cite{shi2022xmorpher}, a unique X-shaped transformer architecture called XMorpher was introduced. The researchers incorporated cross-attention between two feature extraction branches and a window size constraint to enhance information exchange and locality of the network. Another study, Swin-VoxelMorph \cite{zhu2022swin}, utilized a fully Swin Transformer-based 3D Swin-UNet and a bidirectional constraint to optimize both forward and inverse transformation. To fill the gap of affine image registration, Mok \emph{et al.} \cite{mok2022affine} proposed a Coarse-to-Fine Vision Transformer (C2FViT), a pure transformer architecture. The researchers transformed image pairs into small-to-large resolutions and passed them through different stages of ViT to achieve the desired results.

\subsubsection{Analysis}
In CNN, convolution is usually performed within a local region, whereas the Transformer's self-attention mechanism allows for global information exchange. Therefore, integrating the Transformer into the registration network primarily aims to leverage its ability to construct global information, further enhancing the feature extraction. Additionally, there is a growing trend towards designing pure Transformer architectures, demonstrating outstanding performance in visual tasks. However, the attention mechanism should be used in a more registration-specific manner. To address this problem, cross-attention Transformers are being explored at both the feature extraction and feature matching stages. 

It is worth noting that nearly all Transformer-based registration methods are applied in the Brain MRI dataset, so there is no related work on retinal registration. This phenomenon may be caused by access to a more significant number of MR public datasets and an enormous amount of data. At the same time, ViT requires a significantly more significant amount of data to achieve better results than the general CNN network.

\section{Discussion}
\label{sec:discussion}

\subsection{Challenges in Retinal Image Registration}

\subsubsection{Lack of public datasets}
In artificial intelligence, many tasks rely on competition and public evaluation challenges to make progress. These challenges offer a comprehensive and impartial platform for researchers to compare the performance, computation time, and robustness of newly designed algorithms. The Learn2Reg challenge, for example, recently focused on registering medical image modalities commonly used in brain, abdomen, and thorax imaging \cite{L2R}. The currently available datasets for retinal image registration are recorded in Table \ref{dataset_table}. It has not formed enough public retinal image datasets for each modality, nor has there been any competition.

\subsubsection{Lacking deep learning-based linear registration methods}
\begin{figure*}
    \centering
    \subfigure[General Methodology]{
        \begin{minipage}[b]{0.45\linewidth}
            \centering
            \includegraphics[scale=0.45]{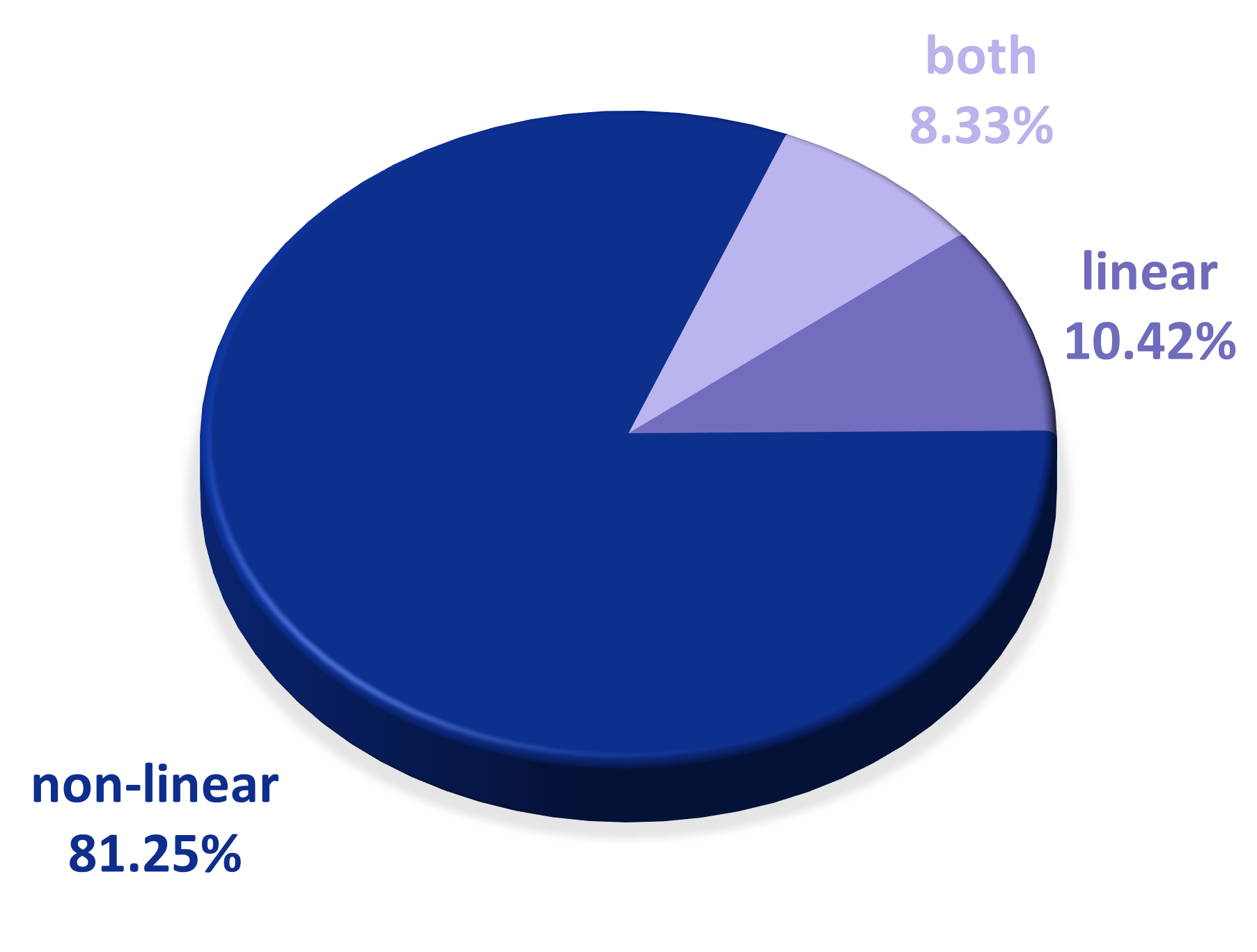}
        \end{minipage}
    }
    \subfigure[Retinal Application]{
        \begin{minipage}[b]{0.45\linewidth}
            \centering
            \includegraphics[scale=0.45]{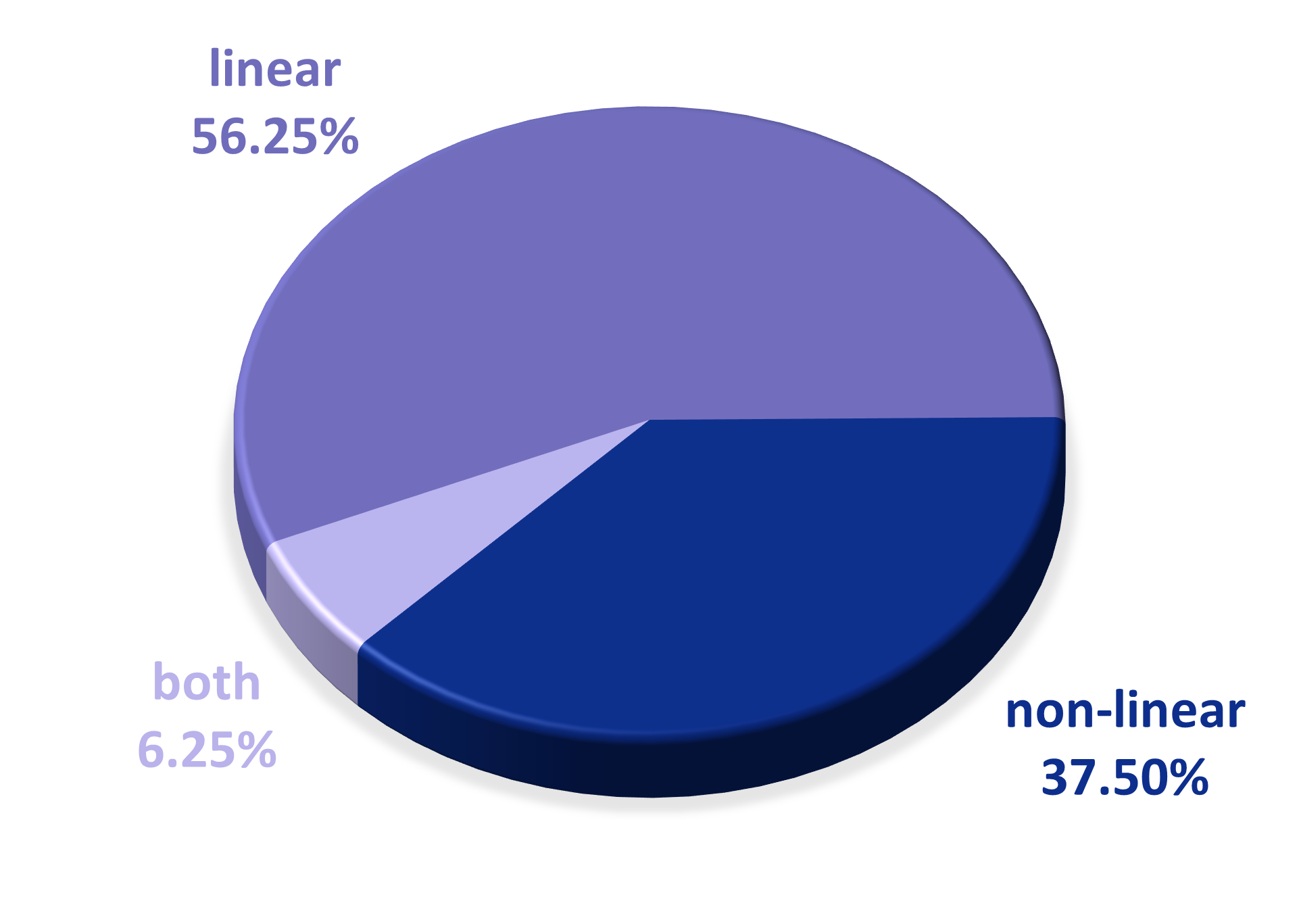}
        \end{minipage}
    }
    \caption{Deep learning-based method using different transformation types}
    \label{tt}
\end{figure*}

Based on the articles using deep learning we reviewed, we calculated the proportion of different transformation types used in general registration methodology and the proportion of each type specifically in the retinal image registration task, as shown in Fig. \ref{tt}. We found that over 80\% of the works on general methodology employ non-linear transformation. However, for retinal applications, linear transformation is the most common method. This is because retinal images are primarily captured from a particular area of the retina, and when registering different images, the contents of these different visual field areas are generally registered. On the other hand, most of the organs captured by commonly used modalities are 3D images and have no difference in field of view. The main differences are in position, inevitable elastic deformation of the organs, or differences caused by different collection objects. Therefore, the registration methods mainly focus on correcting non-linear deformations. Most will directly use traditional algorithms to perform rigid registration for position differences. Because these mainstream registration methods mainly focus on non-linear registration, linear registration tasks for retinal images are not popular.

\subsubsection{Poor similarity metric}
Similar metrics are always used to optimize the registration network in an unsupervised manner or to evaluate the quality of registration. The key technical challenge in medical image registration is selecting and designing the most effective similarity measurement. In uni-modal image registration, brightness changes may present the most significant difficulty. One of the main obstacles in multi-modal image registration is that images from different modalities hold different resolutions, contrast, and luminosity. A newly designed similarity metric or a completely different technical route for multi-modal image registration is urgently needed.

\subsubsection{Intractable retinopathy}
In clinical treatments, most patients suffer from eye retinopathy, so their retina may be severely damaged. Small bulges, swellings, or blood may cover the normal fundus and have a bad effect on photography. Meanwhile, some diseases may change the structure of the retina. Most samples in the public datasets are retinal images of ordinary people. However, when it is used in clinical diagnosis, some patients' retinas are likely to have retinopathy. In this situation, a network trained using normal images cannot work well.

\begin{table*}
    \caption{Public retinal image registration datasets}
    \centering
    \resizebox{\linewidth}{!}{
    \begin{tabular}{p{1.8cm}p{3cm}p{2.5cm}p{1cm}p{1.2cm}lp{1.5cm}p{1.8cm}}
        \toprule
        \textbf{Dataset} & \textbf{Source} & \textbf{Camera Specs.} & \textbf{Format} & \textbf{Modality} & \textbf{Resolution} & \textbf{Size (pairs)} & \textbf{Ground truth}\\
        \midrule
        FIRE \cite{FIRE} & Papageorgiou Hospital, Aristotle University of Thessaloniki, Greece & Nidek AFC-210 fundus camera & JPG & CF & 2912$\times$2912 & 134 & Control Points\\
        FLoRI21 \cite{FLoRI21article} & RECOVERY study \cite{WYKOFF20191076} & Optos California and 200Tx cameras & TIFF & UWF FA & 3900$\times$3072 & 15 & Control Points\\
        CF-FFA \cite{CF-FA} & Unknown & Unknown & JPG & CF \& FFA & 720$\times$576 & 60 & None\\
        \bottomrule
    \end{tabular}
    }
    \label{dataset_table}
\end{table*}

\subsection{Future Scope}
In this era of large models, we may anticipate a general large model for registration to emerge soon. With the ability to use human-marked point pairs or corresponding mask areas as registration prompts, this model will be trained on higher quality, broader types, and more extensive image registration datasets, allowing for better generalization.

There are still plenty of areas to explore regarding retinal image registration. With multiple image modalities, there is a pressing need for multi-modal image registration. To address this, translation-based methods and disentangling representation methods may be another new exploration. Interestingly, we have not seen any pioneers attempting Transformer-based retinal registration methods, which could lead to even greater accuracy.

Moreover, data scarcity remains a significant challenge, but we may overcome it through data generation or transfer learning. For instance, we can supplement the dataset with image pairs obtained through random translation, rotation, and brightness and contrast enhancement using retinal images from other datasets. When employing transfer learning, we can train on endoscopic images from other parts of the human body or manually generate virtual datasets and fine-tune them for retinal image registration.

\section{Conclusion}
\label{sec:conclusion}
This paper thoroughly analyzed medical image registration, focusing on its applications in retinal imaging. Our review compares general medical image registration techniques and their adaptation to retinal imaging, highlights gaps in current research, and gives a little advice on avenues for future exploration. We also evaluated state-of-the-art medical image registration methods and weighed the advantages and disadvantages of each. Lastly, we identified challenges specific to retinal registration and discussed potential opportunities for further advancement.

\backmatter

\section*{Declarations}

\bmhead{Abbreviations}
CAD: Computer-Aided Diagnosis; CNN: Convolutional Neural Network; GAN: Generative Adversarial Network; CF: Color Fundus photography; FA: Fluorescein Angiography; OCT: Optical Coherence Tomography; OCTA: Optical Coherence Tomography Angiography; DR: Diabetic Retinopathy; AMD: Age-related Macular Degeneration; CC: Cross Correlation; MI: Mutual Information; SSD: Sum of Squared Difference; DDPM: Denoising Diffusion Probabilistic Model; ViT: Vision Transformer.

\bmhead{Availability of data and materials}
All relevant data and material are presented in the main paper.

\bmhead{Competing interests}
The authors declare that they have no competing interests.

\bmhead{Funding}
This work was supported in part by General Program of National Natural Science Foundation of China (82102189 and 82272086), and Guangdong Provincial Department of Education (SJZLGC202202).

\bmhead{Authors’ contributions}
All the authors make substantial contribution in this manuscript. QN, YH, and MG participated in writing the first draft of the paper. Then the paper was revised carefully and completed the final manuscript by QN and XZ. JL supervised the study. All the authors have read and approved the final manuscript.

\bmhead{Acknowledgement}
Not applicable.

\bibliography{sn-bibliography}%

\end{document}